\documentclass[12pt]{article}
\usepackage{authblk}
\usepackage{url}
\usepackage{mathptmx}       % selects Times Roman as basic font
\usepackage{helvet}         % selects Helvetica as sans-serif font
\usepackage{courier}        % selects Courier as typewriter font
\usepackage{mathrsfs}
\usepackage{eucal}
\usepackage{mathtools}
\usepackage{makeidx}         % allows index generation
\usepackage{graphicx}        % standard LaTeX graphics tool
\usepackage{multicol}        % used for the two-column index
\usepackage[bottom]{footmisc}% places footnotes at page bottom
\usepackage{esvect}

\newcommand{\thetas}{\mathbf{\theta}}
\newcommand{\etas}{\mathbf{\eta}}

\newcommand{\qs}{\mathbf{Q}}
\newcommand{\hatthetas}{\mathbf{\hat{\theta}}}
\newcommand{\hatetas}{\mathbf{\hat{\eta}}}

\newcommand{\hatqs}{\mathbf{\hat{Q}}}

\DeclareMathOperator*{\argmax}{arg\,max}

\makeindex             % used for the subject index
\begin{document}

\title{Network-based models for social recommender systems}
\author[1]{Antonia Godoy-Lorite}
\author[2,3]{Roger Guimer\`a}
\author[3]{Marta Sales-Pardo}
\affil[1]{Department of Mathematics, Imperial College London, London SW7 2AZ, United Kingdom.}
\affil[2]{Instituci\'o Catalana de Recerca i Estudis Avan\c{c}ats (ICREA), 08010 Barcelona, Catalonia.}
\affil[3]{Departament d'Enginyeria Qu\'{\i}mica, Universitat Rovira i Virgili, 43007 Tarragona, Catalonia.}

\renewcommand\Affilfont{\itshape\small}
\maketitle

\abstract{
With the overwhelming online products available in recent years, there is an increasing need to filter and deliver relevant personalized advice for users. Recommender systems solve this problem by modeling and predicting individual preferences for a great variety of items such as movies, books or research articles. In this chapter, we explore rigorous network-based models that outperform leading approaches for recommendation. The network models  we consider are based on the explicit assumption that there are groups of individuals and of items, and that the preferences of an individual for an item are determined only by their group memberships. The accurate prediction of individual user preferences over items can be accomplished by different methodologies, such as Monte Carlo sampling or Expectation-Maximization methods, the latter resulting in a scalable algorithm which is suitable for large datasets. 
}

\section{Introduction}
\label{sec:Introduction}
The internet has changed the way business is done and how products are advertised, sold and distributed \cite{castells01,kalakota99}. Now we are a click away from an ever increasing array of products \cite{brynjolfsson03}. However, the availability of so many choices puts a stress on the customer who often has no time to browse over endless online product catalogs. As an illustration,  Netflix\index{Netflix} has available around 5,000 movies only in the United States, iTunes more than 37 million songs and Amazon\index{Amazon} up to 32 million books in different formats; if we were to spend 0.5 seconds per item, it would take us approximately 40 minutes to browse the whole catalogue in Netflix\index{Netflix} and over 200 full days  to go through the whole catalogue of songs and books in iTunes and Amazon\index{Amazon}. The platforms that have best adapted to this situation are those that efficiently recommend items that fit personal preferences.

Recommender systems are algorithms precisely designed to predict user's preferences over a variable amount of items. 
A popular event that boosted research in recommender systems was the Netflix\index{Netflix} contest (2006-09) \cite{netflix06,netflix08,netflix09}. Netflix\index{Netflix} sponsored a competition to improve the accuracy of their recommendation algorithm at the time, offering \$1,000,000 to the best performing team. This competition captured the attention of researchers on the topic and improved significantly the state-of-the-art algorithms and even resulted in the creation of companies (for instance, Gravity R\&D or 4-Tell Inc.~\cite{lohr09}) that played a major role in boosting  e-commerce. The increase in the volume of online business coupled to the availability of data on online purchases of products by users, has in recent years enhanced the interest in recommender systems, both in the private and academic sectors.

Currently, the main strategies for making social recommendations are content-based approaches, collaborative filtering, and hybrid  approaches \cite{bobadilla13}. Content-based approaches use available metadata on users or items such as demographics, overall top selling items, past buying habit of users, or item reviews to guess user preferences. On the other hand, collaborative filtering (CF\index{Collaborative Filtering}) methods are based on the plausible expectation that similar users relate to similar objects in a similar manner, i.e., they purchase similar items and rate the same item similarly. Hybrid methods aim at combining both approaches.

Importantly, CF approaches are in general more accurate at predicting user preferences than content-based approaches.
Typically datasets available for recommendation are {\it sparse} -- most users rate just a few items ($ < 10$ items) and most of the items have been rated only by a few users, which makes it hard for content-based methods to make good predictions. In contrast,  CF\index{Collaborative Filtering}\index{Algorithm!Collaborative Filtering} algorithms have successfully addressed this problem by exploiting known preferences of  like-minded users to provide item recommendations or predictions.

A major problem recommender systems face is the need to provide recommendations in a reasonably short amount of time. Taking into account that available datasets comprise millions of  user-item ratings (and that is a small fraction of the data in a real industrial setting), the scalability of the algorithm with the number of observations is critical. Specifically,  if the run-time of an algorithm scales linearly in the number of observed ratings $R$, the time needed to obtain a recommendation if  $R=10,000$  is $\propto 10^5$, whereas if an algorithm scales quadratically with $R$, then the time needed to obtain a recommendation for $R=10,000$ ratings will be $\propto 10^{10}$. As a good rule of thumb, linear (or sub-linear) CF\index{Collaborative Filtering}\index{Algorithm!Collaborative Filtering} algorithms will be good candidates to deal with the large and sparse datasets in an industrial set-up.

In this chapter, we focus on collaborative filtering models suitable for large sparse datasets. Specifically, we show how block model approaches to model a network of  user-item ratings is superior to state-of-the-art approaches such as the Item-Item\index{Item-Item!Algorithm}\index{Algorithm!Item-Item} and Matrix Factorization\index{Matrix Factorization} approache~\cite{deshpande04,sarwar01,koren09,mackey10}.

This chapter is organized as follows. In Sect.~\ref{sec:generalnet} we introduce the network framework for the recommendation problem, fundamental concepts of inference and the use of Stochastic Block Models to make inference on network data. 
In Sect.~\ref{sec:SBM} we present the network-based approaches for recommender systems, the bipartite Stochastic Block Model recommender (SBM) and the Mixed-Membership Stochastic Block Model recommender (MMSBM). In Sect.~\ref{sec:Benchmarcks} we give an overview of some of the most successful collaborative filtering approaches, one being the Item-Item\index{Item-Item!Model}\index{Model!Item-Item} model and two Matrix Factorization\index{Matrix Factorization} approaches: the ``classical'' Matrix Factorization (MF) and the Mixed-Membership Matrix Factorization\index{Mixed-Membership Matrix Factorization}\index{Mixed-Membership!Matrix Factorization}\index{Matrix Factorization!Mixed-Membership}(MMMF), which we use as benchmark algorithms. In Sect.~\ref{sec:Results} we analyze and compare the predictive power of those algorithms and provide a practical guide to run the network-based algorithms with real datasets. To conclude, in Sect.~\ref{sec:discussion} we provide overall discussion of the chapter.

%%%%%%%%%%%%%%%%%%%%%%%%%%%%%%%%%%%%%%%%%%%%%%
\section{Network approach to recommender systems}
\label{sec:generalnet}

Formally, the recommendation problem is the following. We have an observation $R^{\cal O}$ consisting of a collection of ratings $R_{ij}$ of users ($i$) to items ($j$) (eg. ratings of users to  movies or books). Ratings are on a fixed discrete scale $S$, so that each observed rating $r_{ij}$ takes a value within this scale (eg. in a 5 point scale $S=\{ 1,2,3,4,5\}$). This observation is sparse so that out of a group of $N$ users and $M$ items we only observe a small fraction of the $N \times M$ possible ratings. The goal is then to predict the values of a set of query/unobserved ratings $R^{\rm T}$. In the recommender systems literature, the observation  $R^{\cal O}$ is called the training set, while the query set $R^{\rm T}$ is called the test set. This is because recommendation algorithms use the training dataset to train the algorithm and obtain the model parameters and the test set to asses the accuracy of the trained algorithm at making predictions for unobserved ratings. Note that being able to make accurate predictions on unobserved ratings is the first and necessary step toward being able to make suggestions of new items to users based on the rating predictions over those items.

\begin{figure}[!ht]
%\sidecaption[t]
\centering{\includegraphics[width=0.8\columnwidth]{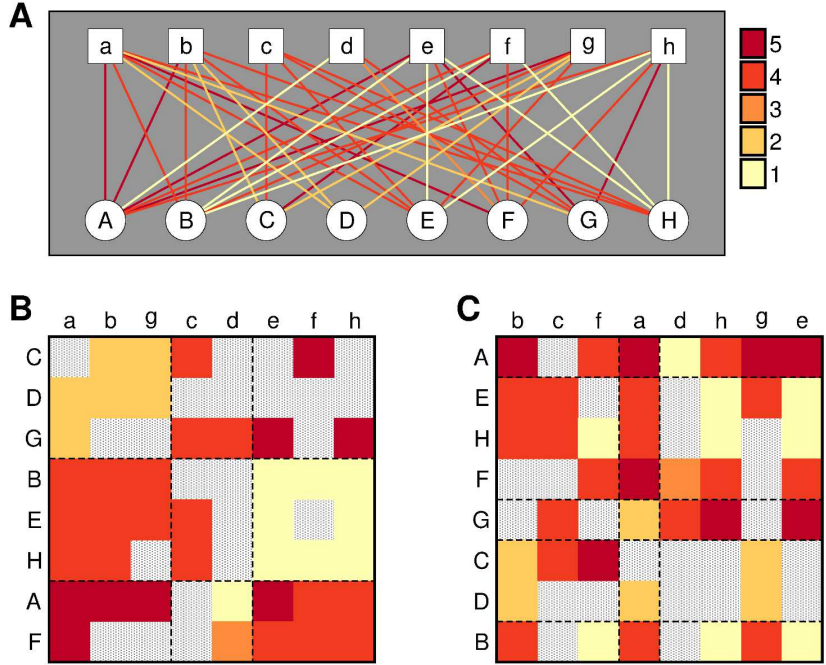}}
%
% If no graphics program available, insert a blank space i.e. use
%\picplace{5cm}{2cm} % Give the correct figure height and width in cm
\caption{{\bf Bipartite SBM for recommendation.} 
	({\bf a}) Eight users llabelled {\sffamily A}--{\sffamily H} rate movies, labelled {\sffamily a}--{\sffamily h}, as indicated by the colors of the links. ({\bf b}-{\bf c}) Matrix representation of the ratings; gray elements represent unobserved ratings. Different partitions of the nodes into groups (indicated by the dashed lines) provide different explanations for the observed ratings. The partition in {\bf b} has a high explanatory power  because ratings in each pair of user-item groups are very homogeneous. For example, it seems plausible that user {\sffamily C} would rate item {\sffamily a} with a 2, given that all users in the same group as {\sffamily C} rate  2 all items in the same group as {\sffamily a}. Conversely, the partition in {\bf c} has very little explanatory power. The predictions of partition {\bf b} contribute much more than those of partition {\bf c} to the inference of unobserved ratings. \textit{Reprinted from Guimer\`a R. et al. Predicting Human Preferences\index{Human Preferences!Prediction}\index{Prediction!Human Preferences} Using the Block Structure\index{Block Structure} of Complex Social Networks. PLOS ONE 7(9):e44620,
	  under the Creative Commons Attribution (CC BY) license at~\protect\url{https://creativecommons.org/licenses/by/4.0/}. }
}
\label{fig:recommender_SBM}      
\end{figure}

Formally, this problem can be mapped into a problem of predicting unobserved  edge values in a network. Specifically, since ratings occur between  two different types of nodes (users and items), we  can  represent  $R^{\cal O}$ as a bipartite network (see Fig.~\ref{fig:recommender_SBM}A). In this network we have an edge connecting each user with all the items she has provided a rating for in the observation. Importantly, within this representation edges can take a value within the scale of ratings $S$. The problem of predicting ratings within the unobserved query set $R^{\rm T}$ then becomes that of predicting values of unobserved edges within this network. 

Here, we focus on estimating the probability that a specific unobserved edge $r_{ij}$ takes value $r$ given the observed data  $R^{\cal O}$, which formally is expressed as $p(r_{ij}=r| R^{\cal O})$. To do that we use a Bayesian\index{Bayesian} approach to perform inference on network data. In what follows, we introduce the basic concepts of Bayesian\index{Bayesian} inference and introduce the Stochastic Block Model, a general class of generative models suitable for network data.

\subsection{Inference on complex networks based on Stochastic Block Models}
\label{sec:Networks}

Let us assume our observed data is $R^{\cal O}$ and we  want to know the probability that a certain variable $X$ (for instance a rating) takes values $x$ conditioned on the observed data, that is $p(X=x|R^{\cal O})$. If we consider an ensemble of generative models $\mathcal{M}$ for our observed data we can express $p(X=x|R^{\cal O})$ as

\begin{equation}
P(X=x|R^{\cal O})=\int_{\mathcal{M}} \: p(X=x|M)\: p(\mathcal{M}|R^{\cal O}) \:dM,
\label{eq:inferece}
\end{equation}
where $p(X=x|M)$ is the probability of variable $X$ being equal to $x$  given model $M$ ($X$ is for 
example the rating that  user $u$ gives item $i$), and $p(M|R^{\cal O})$ is the plausibility of model $M$ given the observation $R^{\cal O}$. Using Bayes' theorem we can rewrite 
Eq.~\ref{eq:inferece} as,

\begin{equation}
P(X|R^{\cal O})=\frac{\int_{\mathcal{M}} \: p(X|M) \: p(R^{\cal O}|M) \: p(M) \: dM}{p(R^{\cal O})},
\label{eq:bayes}
\end{equation}
where $p(R^{\cal O}|M)$ is the probability that model $M$ gives rise to the observed data $R^{\cal O}$, also called likelihood, and $p(M)$ is the prior probability that model $M$ is the correct one, also called the prior. Importantly, the accuracy of the predictions depends strongly on the ability of some of the models in the family of models in $\mathcal{M}$ to describe the observed data.

In our case we consider the  family of Stochastic Block Models (SBM)~\cite{guimera09,guimera12,godoy-lorite16b} as generative models. SBMs are  based on the simple assumption that there are groups of nodes and that nodes within a group have similar connectivity patterns. Within this class of models, the probability of two nodes being connected only depends on the groups to which the nodes belong. Formally, a SBM $\mathcal{M} = (P, \textbf{Q})$ is then completely determined by the partition $P$ of nodes into groups and
the matrix $\textbf{Q}$ of connection probabilities between pairs of nodes belonging to pairs of groups, so that Eq.~\ref{eq:bayes} can be rewritten as
\begin{equation}
P_{\rm SBM}(X=x|R^{\cal O})=\frac{\sum_{P \in \cal{P}}\int_{[0,1]^G} \: p(X=x|P,\textbf{Q}) \: p_{SBM}(R^{\cal O}|P,\textbf{Q}) \: p(P,\textbf{Q}) \: d\textbf{Q}}{p(R^{\cal O})},
\label{eq:bayesSBM}
\end{equation}
where $\cal P$ is the space of all possible partitions of  nodes into groups and $G$ is the total number of pairs of groups of nodes.

\begin{figure}[!ht]
	\centering
	%\sidecaption[t]
	{\includegraphics[width=0.8\columnwidth]{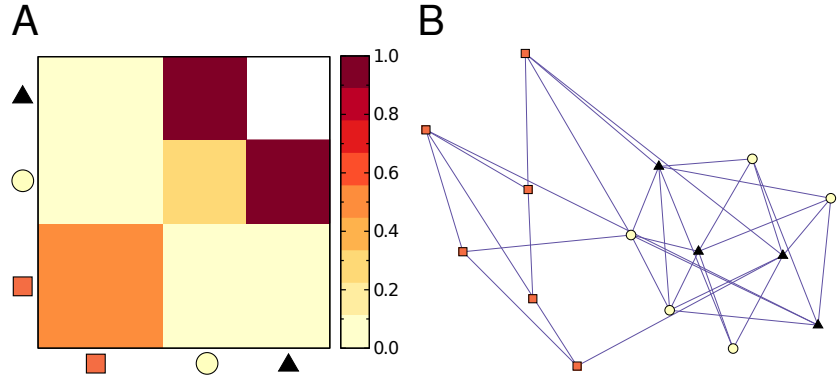}}
	%
	% If no graphics program available, insert a blank space i.e. use
	%\picplace{5cm}{2cm} % Give the correct figure height and width in cm
	\caption{{\bf Stochastic Block Models.} A stochastic block model is fully specified by a partition of nodes into groups $P$ and a connection probability matrix $\textbf{Q}$. Each element $Q_{\alpha \beta}$ in the $\textbf{Q}$ matrix represents the probability that a node in group $\alpha$ connects to a node in group $\beta$.({\bf a}) An example of a $\textbf{Q}$ matrix of connection probabilities. We consider three groups of nodes comprising 4 (triangles), 5 (circles), and 6 (squares) nodes. We color matrix elements  according to their value following the color bar on the right hand side. ({\bf b}) A realization of the model in panel {\bf a}.}
	\label{fig:SBM}       % Give a unique label
\end{figure}

SBMs are suitable models to describe complex networks because they are versatile enough to capture the large variety of connectivity patterns observed in real networks (see Fig. \ref{fig:SBM} for an illustration). For instance, many real-world networks have been found to have a modular or assortative structure in which  nodes within the same group (also called module or community) are more likely to be connected to nodes within the same group than to nodes in other groups~\cite{girvan02,guimera04,guimera07,fortunato10}. A SBM with $Q_{\alpha\alpha}\gg Q_{\alpha\beta}\: \forall \alpha,\beta$ would generate networks with such structure. Interestingly, 
SBMs can also depict other connectivity patterns such as disassortative patterns in which nodes are more likely to connect to nodes in other groups or patterns that define distinct topological roles such as hubs and peripheral nodes in core-periphery structures \cite{rombach14,white76,guimera07}. Importantly, this family of models can be extended to directed \cite{rovira-asenjo13}, and weighted networks~\cite{aicher14,peixoto17}.

\section{Modeling\index{Modeling} ratings using Stochastic Block Models}
\label{sec:SBM}

In this section we will consider two network-based models for recommender systems: the simple bipartite Stochastic Block Model (SBM) \cite{guimera12} and the Mixed-Membership Bipartite Stochastic Block Model\index{Mixed-Membership Bipartite Stochastic Block Model}\index{Model!Mixed-Membership Bipartite Stochastic Block}\index{Model!Bipartite Stochastic Block!Mixed-Membership} (MMSBM) \cite{godoy-lorite16b}. In both models, there are different groups. The difference between these models is that while in the bipartite SBM each user and item belong solely to one group, in the MMSBM users (and items) have a certain probability of belonging to each group of users (items). Importantly, this fact  allows us to describe the network of ratings using fewer groups of users and items and to implement more efficient inference algorithms.

%%%%%%%%%%%%%%%%%%%%%%%%%%%%%%%%%%%%%%%%%%%%%%%%%%%%%%%%%%%%%%%

\subsection{Predictions based on the bipartite SBM recommender system}

Our inference problem is to estimate the probability $p(r_{ui}=r|R^{\cal O})$ that the unobserved rating of item $i$ by user $u$ is $r_{ui}=r$, given the observation $R^{\cal O}$. Hence, by  setting the observable $x$ in Eq.~\ref{eq:bayes} to $X=r_{ui}$ we obtain,
\begin{equation}
p(r_{ui}=r|R^{\cal O})=\frac{\int_{M} \: p(r_{ui}=r|M)\: P(R^{\cal O}|M)\: p(M) \: dM}{\int_{\mathcal{M}'} \: p(R^{\cal O}|M') p(M') \: dM'}~.
\label{eq:rating}
\end{equation}
Where $p(r_{ui}=r|M)$ is the probability that $r_{ui}=r$ if the ratings where actually generated using model $M$, and $p(R^{\cal O}|M)$ is the probability of model $M$ generating the observed ratings $R^{\cal O}$ or likelihood. 

As previously mentioned, we will use the family of Stochastic Block Models $\mathcal{M}_{SBM}$ to describe the observed ratings \cite{white76,nowicki01,guimera09}. In the bipartite SBM users and items are partitioned into two different and independent sets of groups. Therefore, ${\bf Q}$ is a $g_u\times g_i$ rectangular matrix, with $g_u$ and $g_i$ being the number of user and item groups, respectively.  Additionally, because in our network (ratings) edges can take $|S|$ different values, we have one such matrix for each  value of $r$. Therefore, the probability that the rating of user $u$ to item $i$ is equal to $r_{ui}=r$  depends exclusively on the groups $\sigma_u$ and $\sigma_i$ to which user  $u$ and item $i$  respectively belong, so that
\begin{equation}
p(r_{ui}=r)=Q^r_{\sigma_u \sigma_i}.
\end{equation}
Because in the recommender system the possible rating values for a given edge are exclusive, we have the following constraint $\sum_r Q^r_{\sigma_u \sigma_i}=1$ for each (user, item) pair. 

%Therefore, there are a set ratings values $r\in S=\{1,2,3,4,5\}$ ($S$ could take different values as shown in Table~\ref{t-data}), that are considered as independent and non-overlapping types of edges, but the user and item membership, $\sigma_u$ and $\sigma_i$, are common to all ratings. ---I do not understand this!!!

Note that in the bipartite SBM, ratings are considered as independent categories without assuming that the distance between ratings is linear (that is, that $r=3$ is as far from $r=4$, as $r=4$ is from $r=5$).  This poses an advantage over other approaches which assume linearity in the distances between ratings, since users have been found not to perceive equal differences between adjacent ratings (i.e., $r=4$ and $r=5$ might be perceived as closer in rating space than $r=3$ and $r=4$ ~\cite{ekstrand11}). 

Assuming a flat prior over models $p(M)={\rm const.}$, the integral in Eq.~\ref{eq:rating} over all possible values of $Q_{\alpha\beta}$ can be carried out analytically, so that we obtain 
\begin{equation}
p_{SBM}(r_{ui}=r|R^{\cal O})=\frac{1}{{\cal Z}}\sum_{P_U \in {\cal P}_U , P_I \in {\cal P}_I}\left(\frac{n^r_{\sigma_u \sigma_i}+1}{n_{\sigma_u \sigma_i}+|S|}\right)e^{-{\cal H}(P_U,P_I)}
\end{equation} 
where the sum is over all possible partitions of users and items into groups, $n^r_{\sigma_u \sigma_i}$ is the number of ratings with value $r$ observed from users in group $\sigma_u$ to items in group $\sigma_i$, and $n_{\sigma_u \sigma_i}$ is the total number of observed ratings from users in group $\sigma_u$ to items in group $\sigma_i$. The ${\cal H}(P_U,P_I)$ is understood as an energy function or Hamiltonian\index{Hamiltonian!Function}\index{Function!Hamiltonian} which weighs the contribution of each partition of users and items $(P_U,P_I)$ to the sum over all pairs of paritions,
\begin{equation}
{\cal H}( P_U,P_I)=\sum_{\alpha,\beta} \left[ln((n_{\alpha \beta} +|S|-1)!)-\sum_{s=1}^{|S|} ln ((n_{\alpha \beta}^s)!)\right]
\label{eq:hamiltonian}
\end{equation}
and ${\cal Z}=\sum_{(P_U,P_I)} e^{-{\cal H}(P_U,P_I)}$ is called the partition function. In order to estimate the sum over all partitions, \cite{guimera12} estimated $p_{SBM}(r_{ui}=r|R^{\cal O})$ using Metropolis-Hastings sampling \cite{guimera09,jaynes03}. Note that within this approach, no prior assumptions are made on the grouping of the users and items, or in the desired shape of the connection probability matrix, so that the algorithm itself samples/selects those SBMs which provide the best description of the data.

An advantage of this approach is that we obtain the whole distribution for each rating $P_{SBM}(r_{ui}=r|R^{\cal O})$. Therefore, one can choose how to make predictions: using the most likely rating, the mean or the median, among others. In \cite{guimera12}, the authors chose to select the most likely rating
\begin{equation}
r_{ui}=\argmax_{r} \{p_{SBM}(r_{ui}=r|R^{\cal O})\}.
\label{eq:maxrating}
\end{equation}
This probabilistic prediction is in contrast to most  recommender systems like matix factorization and Item-Item\index{Item-Item!Algorithm}\index{Algorithm!Item-Item} algorithms, where the prediction is expressed as a single real number.

The  bipartite SBM recommender we just described has two main advantages: i) it is based on plausible hypotheses about how individuals' preferences arise, and ii) it is mathematically rigorous since it is the result of the full Bayesian probabilistic treatment of the model. However, the correct probabilistic treatment of the model comes at the cost of producing a slow algorithm. The approach above relies on Markov chain\index{Markov Chain}\index{Model!Markov Chain} Monte Carlo sampling  over partitions to make rating predictions, therefore, its computational time does not scale well with the size of the dataset (see Fig.~\ref{fig:allscaling}B). This fact makes it impractical for datasets with millions of ratings \cite{guimera12}. 
%However, for the small datasets with 100k ratings, the bipartite SBM recommender  outperforms state-of-the-art approaches such Item-Item and Matrix Factorization \cite{guimera12} (see also Fig. \ref{fig:compare}).

%%%%%%%%%%%%%%%%%%%%%%%%%%%%%%%%%%%%%%%%%%%%%%%%%%%%%%%%%%%%%%%%%%%%%%%%%%%%%%%%%%%%%%%%%%%%%%%%%%%%%%%%%
\subsection{Predictions based on Mixed-Membership Stochastic Block Model}
\begin{figure}[!ht]
%\sidecaption[t]
\centering{\includegraphics[width=\columnwidth]{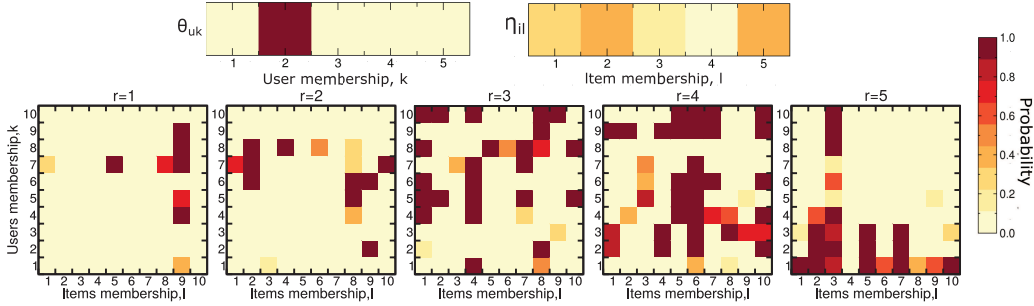}}
%
% If no graphics program available, insert a blank space i.e. use
%\picplace{5cm}{2cm} % Give the correct figure height and width in cm
\caption{{\bf Mixed-Membership Stochastic Block Model.} We illustrate the parameters of a bipartite MMSBM for an equal number of latent groups of users and items $K=L=5$ obtained for  the MovieLens\index{MovieLens} 100K dataset.  In the top row, we show examples of mixed-membership vectors $\thetas$ and $\etas$ for user $u$  and item $i$, respectively. Each vector component $\theta_{uk}$ ($\eta_{il}$) is the probability that user $u$ (item $i$) belongs to group $k$ (group $l$). Probabilities are shown in colors following the color bar on the right hand side.  In this example, user $u$ has a higher probability to belong to group $k=2$, while item $i$ has similar probabilities to belong to any group. In the bottom row, we show the inferred values for the probability matrices $\qs$.  From left to right, the five matrices correspond to the ratings $r=1,2,3,4,5$.  For each one of these matrices, the rows and columns correspond to  user and item groups, respectively.  Each matrix element is the probability $Q^r_{k\ell}$ that a user in group $k$ gives a rating $r$ to an item in group $\ell$. Notice that there is no ordering of the probability matrices that would make them diagonal.
%Original source:  \protect\cite{guimera09}. \textit{Reprinted with permission.}.
}
\label{fig:recommeder_MMSBM}      
\end{figure}

In this section, the Mixed-Membership Bipartite Stochastic Block Model\index{Mixed-Membership Bipartite Stochastic Block Model}\index{Model!Mixed-Membership Bipartite Stochastic Block}\index{Model!Bipartite Stochastic Block!Mixed-Membership} (MMSBM) approach for recommendation is considered. As previously mentioned, mixed membership models allow nodes to belong to all possible (latent) groups with a 
finite probability~\cite{airoldi08,ball-karrer-newman11}. In our case, we consider a bipartite MMSBM in which we have latent groups for  it assumes that each node in the bipartite graph of users and items belongs to a mixture of groups. 

In the recommendation problem our goal is  to estimate the probability $p(r_{ui}=r|R^{\cal O})$. In order to do so, we need to compute the likelihood of the observed data given the model parameters. To that end, we define the model parameters as follows.

In the bipartite MMSBM, we consider that there are $K$ groups of users and $L$ groups of items.  For each pair of user-item groups $k,\ell$, there is a probability $Q^r_{k\ell}\; \forall r \in S$ that users in group $k$  give rating $r$ to items in group $\ell$. Note that because in $R^{\cal O}$ each user-item edge has only one rating $r$ the probability matrices $Q^r_{k\ell}$ are normalized 
\begin{equation}
\label{eq:norm-pr}
\forall k,\ell: \sum_{r \in S} Q^r_{k\ell} = 1 \, .
\end{equation}

To model mixed group memberships, each user $u$ has a vector $\theta_u \in R^K$, where $\theta_{uk}$ denotes the extent to which user $u$ belongs to group $k$.  Similarly, each item $i$ has a vector $\eta_i \in R^L$ (see Fig.~\ref{fig:recommeder_MMSBM}). These vectors are normalized as, 
\begin{align}
\sum_k \theta_{uk} = 1,\\
\sum_\ell \eta_{i\ell} = 1.
\label{eq:norm-pm}
\end{align}

Given the membership vectors $\theta_u$ and $\eta_i$, and the probability matrices $Q^r_{k\ell}$, the probability distribution of each rating $r_{ui}$ is a convex combination,
\begin{equation}
 p(r_{ui} = r) = \sum_{k,\ell} \theta_{uk} \eta_{i\ell} Q^r_{k\ell} \, .
 \label{eq:model}
\end{equation}

Abreviating all these parameters as $\thetas, \etas, \qs$, the likelihood of the observed ratings is thus 
\begin{equation}
P(R^{\cal O} | \thetas, \etas, \qs) = \prod_{(u,i) \in R^{\cal O}} \sum_{k,\ell} \theta_{uk} \eta_{i\ell} Q^{r_{ui}}_{k\ell} \, . 
\label{eq:likeli}
\end{equation}

In order to perform a full Bayesian\index{Bayesian} approach as for the simple bipartite SBM, we would have to compute the integral in Eq.~\ref{eq:bayes} to obtain $p(r_{ui}=r|R^{\cal O})$.
However, this is unfeasible for the current model. Therefore, in order to estimate $p(r_{ui}=r|R^{\cal O})$, we make a steepest descent approximation and evaluate the integral by considering the model parameters $\hatthetas, \hatetas$, and $\hatqs$  that maximize the likelihood in Eq.~\ref{eq:likeli}.

%However, the MMSBM approach in \cite{godoy-lorite16} used a maximum likelihood estimate of the the Bayesian inference . That is, instead of integrate over all possible models, the authors evaluate the integral at the maximum likelihood point as,

%\begin{equation}
%p(r_{ui}= r|R^{\cal O})\simeq \left. \frac{\int_{\mathcal{M}} \: d\mathcal{M} \: p(r_{ui}=r|\mathcal{M})\: P(R^{\cal O}|M)\: p(\mathcal{M})}{\int_{\mathcal{M}'} \: d\mathcal{M}' \: p(R^0|\mathcal{M}') p(\mathcal{M}')}\right|_{\hat{M}}
%\label{eq:rating}
%\end{equation}
%Where $\hat{M}$ is the model with higher probability given $R^{\cal O}$. Then the inference problem turn into a maximum likelihood problem. That is, from a family model $M(\mathbf{\thetas})$, to find the model parameters $\hatthetas$ that maximized the likelihood function $p(R^{\cal O}|\mathcal{M}(\thetas))$. 

Note that while this approximation should in principle not perform as well as considering all possible model parameters, our results show that the maximum likelihood prediction for the bipartite MMSBM produces as accurate predictions as the full probabilistic treatment of the simple bipartite SBM (Fig.~\ref{fig:compare}). Our results suggest  that the introduction of mixed-membership vectors seems to already  provide enough flexibility to the model to capture all the patterns covered by the model averaging in the simple bipartite SBM approach.
The maximum likelihood parameters $\hatthetas, \hatetas, \hatqs$ are inferred using an efficient expectation-maximization algorithm (EM).  We start with a standard variational trick that changes the log of a sum into a sum of logs, writing
\begin{align}
\log P(R^{\cal O} | \thetas, \etas, \qs ) 
&= \sum_{(u,i) \in R^{\cal O}} \log \sum_{k\ell} \theta_{uk} \eta_{i\ell} Q^{r_{ui}}_{k\ell} \nonumber \\
&= \sum_{(u,i) \in R^{\cal O}} \log \sum_{k\ell} \omega_{ui}(k,\ell) \,\frac{\theta_{uk} \eta_{i\ell} Q^{r_{ui}}_{k\ell}}{\omega_{ui}(k,\ell)} 
\nonumber \\
&\ge \sum_{(u,i) \in R^{\cal O}} \sum_{k\ell} \omega_{ui}(k,\ell) \log \frac{\theta_{uk} \eta_{i\ell} Q^{r_{ui}}_{k\ell}}{\omega_{ui}(k,\ell)} \, . 
\label{eq:finalL}
\end{align}
Here $\omega_{ui}(k,\ell)$ is the estimated probability that a given ranking $r_{ui}$ is due to $u$ and $i$ belonging to groups $k$ and $\ell$ respectively, and the lower bound in the third line is Jensen's inequality $\log \bar{x} \ge \overline{\log x}$. The equality holds when 
\begin{equation}
\label{eq:update-omega}
\omega_{ui}(k,\ell) 
= \frac{\theta_{uk} \eta_{i\ell} Q^{r_{ui}}_{k\ell}}{\sum_{k'\ell'} \theta_{uk'} \eta_{i \ell'} Q^{r_{ui}}_{k'\ell'}} \, , 
\end{equation}
giving us the update Eq.~\eqref{eq:update-omega} for the expectation step.
For the maximization step, we derive update equations for the parameters $\thetas, \etas, \qs$ by taking derivatives of the log-likelihood~\eqref{eq:finalL}.  Including Lagrange\index{Lagrange!Multiplier} multipliers for the normalization constraints~\eqref{eq:norm-pm}, we obtain 
\begin{equation}
\label{eq:update-theta}
\theta_{uk} = \frac{\sum_{i \in \partial u} \sum_l \omega_{ui}(k,\ell)}
{\sum_{i \in \partial u} \sum_{k\ell} \omega_{ui}(k,\ell)} 
= \frac{\sum_{i \in \partial u} \sum_l \omega_{ui}(k,\ell)}{d_u} \, ,
%\mid (u,i) \in R 
\end{equation}
where $d_u$ is the degree of the user $u$.  Similarly,
\begin{equation}
\label{eq:update-eta}
\eta_{i\ell} = \frac
{\sum_{u \in \partial i} \sum_k \omega_{ui}(k,\ell)}
{\sum_{u \in \partial i} \sum_{k\ell} \omega_{ui}(k,\ell)} 
= \frac{\sum_{u \in \partial i} \sum_k \omega_{ui}(k,\ell)}{d_i} \, ,
\end{equation}
where $d_i$ is the degree of item $i$. Finally, including a 
Lagrange\index{Lagrange!Multiplier} multiplier for~\eqref{eq:norm-pr}, we have

\begin{equation}
\label{eq:update-pr}
Q^{r}_{k\ell} = \frac
{\sum_{(u,i) \in R^{\cal O} | r_{ui}=r} \omega_{ui}(k,\ell)}
{\sum_{(u,i) \in R^{\cal O}} \omega_{ui}(k,\ell)} \, .
\end{equation}
These equations can be solved iteratively with the following EM algorithm. Starting with an initial estimate of $\thetas$, $\etas$, and $\qs$, we repeat the following steps until the parameters converge:
\begin{enumerate}
\item (Expectation step) use~\eqref{eq:update-omega} to compute $\omega_{ui}(k,\ell)$ for $(u,i) \in R^{\cal O}$,
\item (Maximization step) use~\eqref{eq:update-theta}-\eqref{eq:update-pr} to compute $\thetas$, $\etas$, and $\qs$.
\end{enumerate}
The number of parameters and terms in the sums in Eqs.~\eqref{eq:update-omega}-\eqref{eq:update-pr} is $NK+ML+|R^{\cal O}| KL$.  Assuming that $K$ and $L$ are constant, this is $O(N+M+|R^{\cal O}|)$, and hence linear in the size of the observed ratings (see Fig.~\ref{fig:allscaling}A).  As the set of observed ratings $R^{\cal O}$ is typically very sparse because only a small fraction of all possible user-item pairs have observed ratings, the expectation-maximization algorithm is feasible even for very large datasets. 

In summary, the MMSBM approach has a double advantage: (i) it uses a model that is realistic and flexible, and (ii) the algorithm scales with the number of observed ratings, and is therefore suitable for very large datasets. In addition, it is consistent for sparse datasets, giving good results with few ratings per user (users in datasets in Sect.~\ref{sec:Results} rate typically less than 10 items, but they are enough to give good predictions).

\section{State of the art: other non-network based collaborative filtering approaches.} 
\label{sec:Benchmarcks}
%Other non-network based collaborative filtering approaches:
%We shall now take up a significantly different CF approaches for social recommendation.
As already mentioned, collaborative filtering algorithms find similarities between users and items to make predictions, instead of focusing on the content or known external information regarding users or items other than user-item ratings. There are different strategies to identify these similarities or patterns in the recommender system. Two of the most representative approaches are neighbor-based  models such as Item-Item\index{Item-Item!Algorithm}\index{Algorithm!Item-Item} or User-User\index{User-User Algorithm}\index{Algorithm!User-User} approaches, and latent factor models such as Matrix Factorization\index{Matrix Factorization}, commonly used also as benchmark algorithms to compare against novel recommendation models.  While neighbor-based models are simple and intuitive, Matrix Factorization\index{Matrix Factorization} techniques are usually more effective because they allow us to discover the latent features underlying the interactions between users and items. Neighbor-based models are sometimes considered graph-based models, given that they use the structure of the bipartite network to compute similarities between users or between items, and they have also been called model-based algorithms \cite{sarwar01}. In this chapter, we will consider as network-based models only those using network inference. Within this section, we explain the rationale for some of the most widely used CF\index{Collaborative Filtering}\index{Algorithm!Collaborative Filtering} algorithms and analyze some of the main theoretical differences between them and the network-based models SBM and MMSBM.

\subparagraph{Item-Item\index{Item-Item!Algorithm}\index{Algorithm!Item-Item}}

Neighbor-based CF\index{Neighbor-based CF}\index{Collaborative Filtering!Neighbor-based}\index{Model!Neighbor-based Collaborative Filtering} models generate recommendations using only information about rating profiles for different users. There are two approaches, the User-User\index{User-User Algorithm}\index{Algorithm!User-User} approach and the Item-Item approach. In the former, the algorithm finds users with a rating history similar to the query user (neighbors) and generates recommendations using this neighborhood; for the Item-Item\index{Item-Item!Algorithm}\index{Algorithm!Item-Item}, the algorithm finds similar items to the query item based on their rating history, and generates recommendations using the query item's neighborhood. For rating systems with much more users than items (as is the case in the datasets we analyze), the Item-Item\index{Item-Item!Algorithm}\index{Algorithm!Item-Item} approach gives better predictions than the User-User\index{User-User Algorithm}\index{Algorithm!User-User} approach and is computationally more efficient, therefore from now on we will focus on the Item-Item\index{Item-Item!Algorithm}\index{Algorithm!Item-Item} algorithm, taking into account that the User-User\index{User-User Model}\index{Model!User-User} model is computed analogously \cite{sarwar01}. Let us assume that we have a list of users $U=\{u_1,...,u_N\}$ and items $I=\{i_1,...,i_M\}$, which the users have rated. The Item-Item approach assumes that the rating from user $u$ to an item $i$ should be similar to the rating she gives to similar items to $i$. Considering the vector $\vv{i} \in R^N$ of ones for users that have rated item $i$ and zeros otherwise, we can obtain the similarity between item pairs $(i,j)$ by computing the cosine similarity between $\vv{i}$ and $\vv{j}$ as, 
\begin{equation}
\mathrm{sim}(i,j)=\cos(i,j)=\frac{\vv{i} \cdot \vv{j}}{|| \vv{i} ||_2  || \vv{j} ||_2},
\label{eq:itemsim}
\end{equation}
where $|| \cdot ||_2$ denotes the Euclidean norm of a vector. For other adjusted versions of the similarity see \cite{sarwar01}. Note that we can only establish similarities between items that have been rated by the same users \footnote{For the User-User\index{User-User Model}\index{Model!User-User} model the reasoning is equivalent: each user is represented by a vector with all the items she has rated $V_u$. The similarity between users would be computed as in equation \ref{eq:itemsim} as $\mathrm{sim}(u,v)$}.  According to the similarity measure, we define the neighborhood of an item $i$, $\partial i$ as those $k$ items with highest similarity $i$. Hence, the prediction of $r_{ui}$ would be the average of the ratings that user $u$ gave to the items in the neighborhood of item $i$ as,
%for the predictions we define thethe $k$ measuring the similarity of the items in the system, and taken the $k$ more similar items, the predicted ratings $r_{ui}$ would be the average of the ratings of users $u$ on the $k$ most similar items,
\begin{equation}
r_{ui}=\frac{\sum_{j\in \partial i}(\mathrm{sim}(i,j)\cdot r_{uj})}{\sum_{j\in \partial i} (|\mathrm{sim}(i,j)|)}.
\end{equation}

Note that if in the k-nearest neighbors there is no item rated by $u$, the algorithm cannot perform a prediction, which may happen for sparse datasets. Also, the algorithm assumes a linear psychological scale on the ratings, that is a rating of 5 is seen as five times better than a rating of 1, but unfortunately this is not necessary in agreement with people's perception~\cite{ekstrand11}.

\subparagraph{Matrix Factorization approaches}

The most widely used methods for recommendations are the Latent\index{Latent!Feature} feature or Matrix Factorization\index{Matrix Factorization} methods \cite{koren09,paterek07}.
% The intuition behind using MF to solve a recommendation problem is that there should be some latent features that determine how a user rates an item.
%The matrix factorization method based on singular value decomposition (SVD) works as follows \cite{koren09}. 
Latent\index{Latent!Feature} feature models assume that there is a space of latent attributes of users and items that determine user-item ratings. Therefore, ratings are not independent from one another but set by the specific position in the latent feature space of users and items. Specifically, MF assumes that there exists a single latent feature space for both users and items, and that the rating of user $u$ to item $i$ is proportional to the closeness between the two in this space. The dimension of the latent space $K$ is much smaller than the number of users and items, such that the problem is dimensionally reduced. 
Formally, this is equivalent to assuming that the matrix of observed ratings $R^{\cal O}$ (with a number of rows equal to the number of users $N$, and a number of columns equal to the number of items $M$) can be decomposed into
\begin{equation}
R^{\cal O}=P\: Q,
\label{eq:MFmatrius}
\end{equation}
where $P$ is a $N\: \times \: K$ matrix associated with the users and $Q$ is a $K\: \times\: M$ matrix associated with the items. Each row of the $P$ matrix $p_u$ could be seen as a $K$-dimensional vector with the feature values of user $u$ that describe her, and each column of the $Q$ matrix $q_i$ is a K-dimensional vector with the values of the features that describe item $i$, with $K<<N$ and $K<<M$.

\begin{figure}[!ht]
%\sidecaption[t]
\centering
\includegraphics[width=\columnwidth]{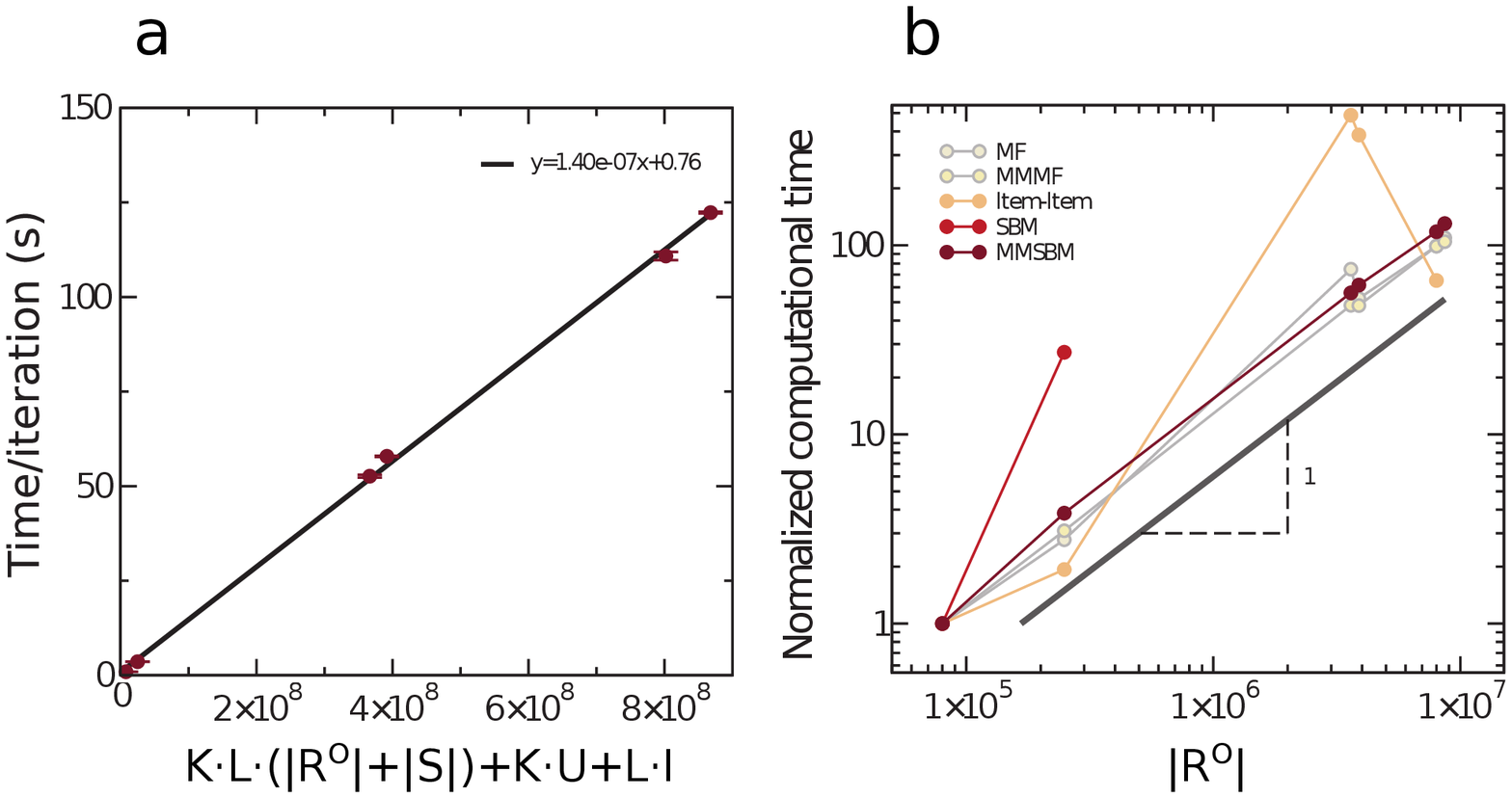}
% If no graphics program available, insert a blank space i.e. use
%\picplace{5cm}{2cm} % Give the correct figure height and width in cm
\caption{{\bf Scalability\index{Scalability}.} ({\bf a}) Scalability of the MMSBM algorithm. Each point represents the average time per iteration in seconds for each of the datasets we use in the study (100K MovieLens, 10M MovieLens\index{MovieLens}, Yahoo! Songs\index{Yahoo! Songs}\index{Dataset!Yahoo Songs}, W-M dating agency, M-W dating agency and Amazon books\index{Amazon!Books Dataset}\index{Dataset!Amazon Books}) versus the number of parameters computed at each iteration $K*L*(|R^{\cal O}|+|S|)+K*N+L*M$ where $N$ is the number of users, $M$ is the number of items and  $|R^{\cal O}|$ is the number of observed ratings,  $|S|$ is the number of different ratings values for each recommender systems and $K$ and $L$ are the number of groups for users and items, respectively ($K=L=10$ for all the datasets; see Table 1 for remaining parameters for each dataset). The continuous line is the linear fit of the real data, which shows that the computational times per iteration scales linearly with the size of the corpus for the whole range.  ({\bf b}) Scaling\index{Scaling} of the different benchmark algorithms we consider in our analysis with the total number of observed ratings. The vertical axis is normalized by the computational time of the smallest dataset -- 100K MovieLens\index{MovieLens}. MF, MMMF and MMSBM algorithms scale linearly with the total number of observed edges, while the Item-Item\index{Item-Item!Algorithm}\index{Algorithm!Item-Item} algorithm does not. Note that for the bipartite SBM we could only get results for the two smallest datasets, so we cannot establish a linear relationship in this case. 
}
\label{fig:allscaling}      
\end{figure}

The most efficient method until now to factorize the rating matrix, although there are several methods, is the singular value decomposition (SVD) \cite{paterek07}. This method finds the two smaller matrices whose product minimizes the difference with the original ratings matrix (measured as a means squared error). In addition it uses gradient descent to learn a Matrix Factorization\index{Matrix Factorization} (by taking derivatives of the error function over the parameters of the model it is trying to infer). The predicted rating is then
\begin{equation}
r_{ui}=\sum_{k} p_{uk} q_{ik}.
\label{eq:MFrating}
\end{equation}
SVD-MF algorithm is computationally very efficient and makes very good predictions. Also, it has the advantage that it results in intuitive meanings of the resultant matrices and that the resulting algorithm is scalable (see Fig.~\ref{fig:allscaling}) so it can potentially handle very large datasets.
The main problem with this approach is that features that describe the users and the items are the same, which imposes severe constraints on the expressiveness of the model (for instance, two users close in feature space must like the same type of items and there is little flexibility to account for the fact that some users might like some items but have different opinions about other items).

Moreover, as the prediction is (with some corrections) the scalar product of the users' and items' feature vectors, this is equivalent to assuming linearity between ratings insted of assuming that ratings are independent categories as was assumed for the bipartite SBM and MMSBM. 

As an extension to the ``classical'' MF, we also consider a mixed-membership implementation of MF, the Mixed-Membership Matrix Factorization\index{Mixed-Membership Matrix Factorization}\index{Mixed-Membership!Matrix Factorization}\index{Matrix Factorization!Mixed-Membership} (MMMF)~\cite{mackey10}. The MMMF model combines Matrix Factorization\index{Matrix Factorization} with a mixed membership context bias. In MMMF, users and
 items are endowed with both latent factor vectors ($p_u$
 and $q_i$) and discrete topic distribution parameters ($\theta_{uk}^U \in K^U$ and $\theta_{ij}^M \in K^M$ ). Together with the user and item topics, MMMF models also introduce the affinity of user $u$ to item topic $k$ as $c_{u}^k$ and the affinity of item
 $i$ to user topic $j$ as $d_{i}^j$. The topic distribution parameters and the affinity of users and items to the topics jointly specify a context bias $\beta_{ui}^{jk}$. Therefore, a user generates a rating for an item by adding the contextual bias to the MF inner product with some Gaussian\index{Gaussian!Noise} noise,
\begin{equation}
r_{ui} \sim \mathcal{N}( p_u \cdot q_i + \beta_{ui}^{jk},\sigma^2).
\label{gauss}
\end{equation}
In~\cite{mackey10} authors consider two different MMMF models that differ in how the contextual bias is built. The Topic-Indexed\index{Topic-Indexed Bias Model}\index{Model!Topic-Indexed Bias} Bias Model (TIB) assumes that the contextual bias decomposes into a latent user
bias and a latent item bias so that $\beta_{ui}^{jk}=\sum_{k=1}^{K^M}c_u^{t(t)} \theta_{ik}^{M(y)} +\sum_{j=1}^{K^U} d_i^{j(t)}\theta_{uj}^{U(t)}$. The Topic-Indexed factor Model (TIF) assumes that the joint contextual bias is an inner product of topic-indexed factor vectors, so that $\beta_{ui}^{jk}=\sum_{k=1}^{K^M} \sum_{j=1}^{K^U} \theta_{ik}^{M(y)} \theta_{uj}^{U(t)} c_u^{k(t)} \cdot  d_i^{j(t)}$. They use a Gibbs sampling MCMC procedure to draw samples of topic and parameter variables. Then, the posterior mean prediction for each user-item pair under these  MMMF models is,
\begin{equation}
\frac{1}{T}\sum_{t=1}^T\left(p_u^{(t)} \cdot q_i^{(t)} + \beta_{ui}^{jk} \right).
\end{equation}
The results shown in Sect.~\ref{sec:Results} are for the MMMF-TIF model since it outperforms the MMMF-TIB in all the datasets. Note that analogously to the MF, MMMF also assumes linearity between ratings values.

\subsection{Advantages of network-based models}

There are a number of advantages to using the network-based models we have presented (the bipartite SBM and the MMSBM) compared to previous work on collaborative filtering. 

First, unlike Matrix Factorization\index{Matrix Factorization} approaches such as~\cite{koren09} or their probabilistic counterparts~\cite{meeds06,salakhutdinov08,shan10}, the ratings $r_{ui} \in \{1,2,3,4,5\}$ are not treated as integers. As has been established in the literature, giving a movie a rating of 5 instead of 1 does not mean the user likes it five times more~\cite{ekstrand11}. Indeed, the results in Sect.~\ref{sec:Results} suggest that it is better to think of different ratings simply as different labels on the links of the network. 

Second, network-based methods yield a distribution over the possible ratings directly, rather than a distribution over integers or reals that must be somehow mapped to the space of possible ratings~\cite{meeds06,salakhutdinov08,shan10}. The network-based  models we have presented considered the observed ratings as a bipartite network with metadata (or labels) on the links. An alternative approach would be to consider a multi-layer representation of the data as in~\cite{peixoto15}.

Third, the bipartite SBM and the MMSBM do not assume that the matrices $\qs$ have any particular structure. In particular, they do not assume either that groups of individuals correspond to groups of items, or that individuals prefer items that belong to their own group (which mathematically would result in diagonal $\qs$ matrices). Thus, the SBMs and the resulting algorithms, can learn arbitrary couplings between groups of individuals and groups of items, and do so independently for each possible rating, thus overcoming the limitation of expressivity of MF factorization approaches that consider a diagonal $\qs$ matrix. Importantly, the MMMF does not circumvent this issue despite considering arbitrary couplings between users/items and topics. In fact, MMMF rating predictions are the sum of a MF term and a correction that uses mixed group memberships that are unrelated to the feature vectors~\cite{mackey10}.  While this is an improvement over MF, it does not fundamentally remove the limiting assumption that each group of users has a corresponding group of items that they prefer.  Indeed, our numerical results show that the performance of MMMF is fairly close to that of MF in the datasets we considered. 

Finally, all the network-based models presented here do not assume that individuals only see movies (say) that they like, and they do not treat missing links as zeroes or low ratings as is typically done in MF algorithms that need a full matrix to decompose. There are other physics-inspired methods that exploit the structure of the bipartite user-item network and use classical physics processes to make recommendations such as random walks~\cite{zhou07} or heat diffusion~\cite{zhou10}. However, all these approaches are used for link prediction, that is, they only try to predict which item would be collected by a user~\cite{yu16}.
% To put this differently, they only try to predict the unobserved ratings in $R \setminus R^{\cal O}$. 

%Now regarding the differences between the bipartite SBM and the MMSBM, the mixing of communities of the MMSBM has the advantage of being mathematically tractable. It yields an expectation-maximization algorithm for fitting the parameters which is highly efficient: each iteration takes linear time as a function of the number of users, items, and observed links.  As a result, it is able to handle quite large datasets, and achieve a higher accuracy than standard methods.

%
\section{Results}
\label{sec:Results}
We show the performance of the network-based and the Item-Item\index{Item-Item!Algorithm}\index{Algorithm!Item-Item} and MF algorithms for six datasets: the MovieLens\index{MovieLens} 100K and 10M datasets with 100,000 and 10,000,000 ratings, respectively (\url{https://movielens.org}), Yahoo! songs (R3--\url{https://webscope.sandbox.yahoo.com/catalog.php?datatype=r&guccounter=1}),  Amazon books\index{Amazon!Books Dataset}\index{Dataset!Amazon Books}~\cite{mcAuley15,mcAuley15a} (\url{http://jmcauley.ucsd.edu/data/amazon/}), and the dataset from LibimSeTi.cz dating agency~\cite{brozovsky07} (\url{http://www.occamslab.com/petricek/data/}), which is split into two datasets, one consisting of males rating females and vice versa. These datasets are diverse in the types of items considered, the sizes $|S|$ of the sets of possible ratings, and the density of observed ratings (see Table~\ref{t-data}).

\begin{table}
\caption{{\bf Dataset characteristics.} The total number of possible ratings is different for each dataset; ratings are in a scale from 1 to 5 in all datasets for the two dating agency datasets, which have a rating scale from 1 to 10.  Ratings are integers except for the Movielens 10M dataset which allows half-integer values.  Note that, in the latter case we expect a smaller MAE than if only integer values were allowed.  
%we had the 10 ratings in a scale from 1 to 10 with a precision of 1.  
All datasets have millions of ratings except for MovieLens\index{MovieLens} 100K and Yahoo! Songs\index{Yahoo! Songs}\index{Dataset!Yahoo Songs}.}  
\begin{center}
\begin{tabular}{@{\vrule height 10.5pt depth0pt  width0pt}l|c|r|r|r}
\hline
Dataset &Ratings scale $S$&\#Users &\#Items &\#Ratings  \\ 
%& & & & & cold start (\%)\\
\hline
MovieLens 100K &$\{1,2,3,4,5\}$ &943 & 1,682 & 100,000 \\
%MovieLens& & & & & \\
MovieLens 10M & $\{0.5,1, 1.5, \ldots ,5\}$& 71,567 & 65,133 & 10,000,000 \\
%MovieLens& & & & & \\
Yahoo! Songs\index{Yahoo! Songs}\index{Dataset!Yahoo Songs} & $\{1,2,3,4,5\}$  & 15,400 & 1,000 & 311,700 \\
M-W dating agency&$\{1,2, \ldots,10\}$ & 220,970  & 135,359 & 4,852,455 \\
%dating agency& & & & & \\
W-M dating agency&$\{1,2, \ldots,10\}$ & 135,359 &220,970 & 10,804,040 \\
%dating agency& & & & & \\
Amazon book & $\{1,2,3,4,5\}$ & 73,091  & 539,145 & 4,505,893 \\
\hline
\end{tabular}
\end{center}
\label{t-data}
\end{table}

\begin{figure}[!ht]
%\sidecaption[t]
\centering{\includegraphics[width=0.635\columnwidth]{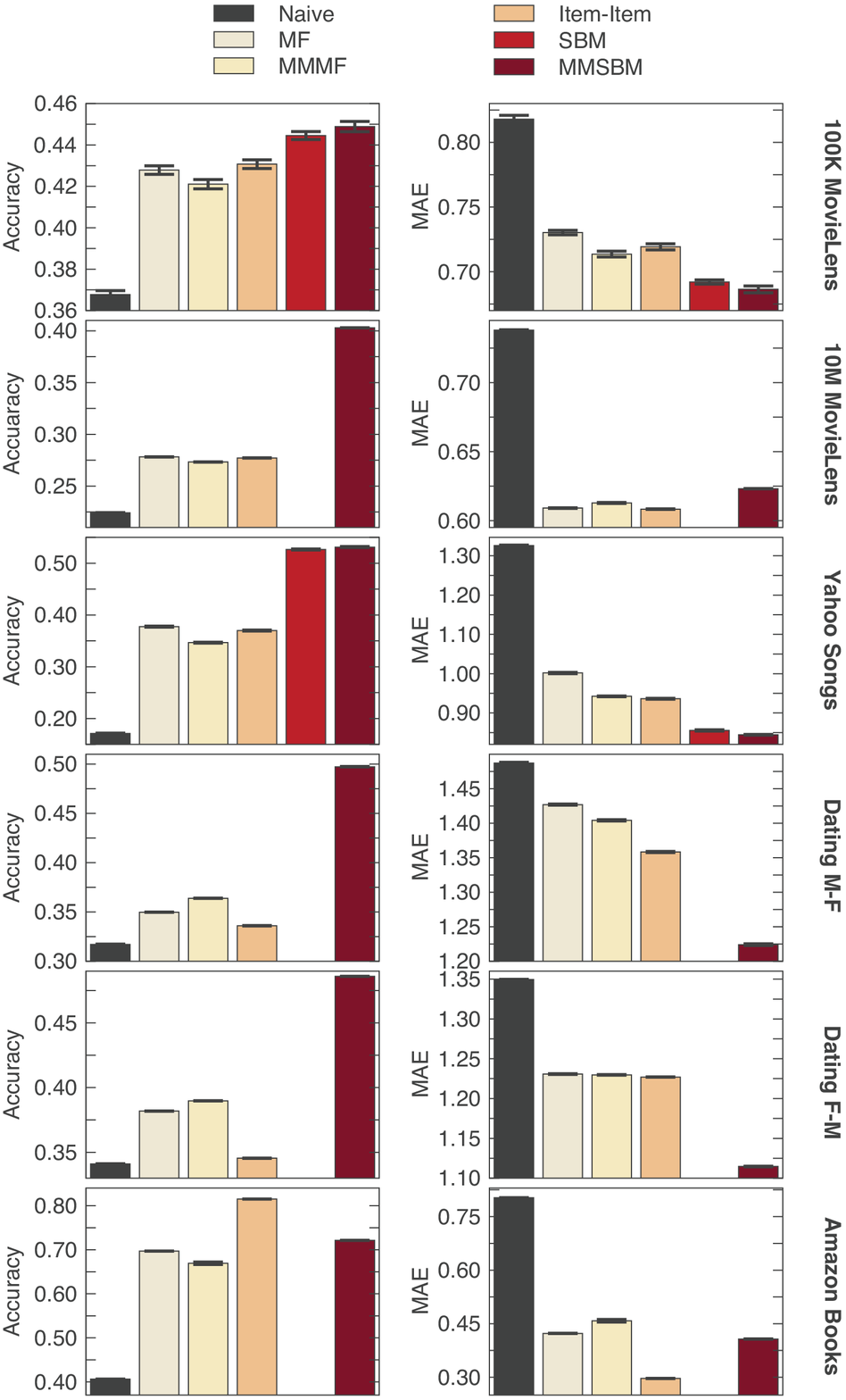}}
%
% If no graphics program available, insert a blank space i.e. use
%\picplace{5cm}{2cm} % Give the correct figure height and width in cm
\caption{{\bf Algorithm comparison.} From top to bottom, the datasets are
MovieLens 100K,  % (with ratings $S=\{1,\ldots,5\}$), 
Movielens 10M, %(where $S=\{0.5,1,1.5,\ldots,5\}$), 
Yahoo Songs,  %(311,700 ratings, $S=\{1,\ldots,5\}$), 
men rating women (M-W) in the LibimSeTi dataset, % (4,852,455 ratings, $S=\{1,\ldots,10\}$), 
women rating men (W-M)  in the LibimSeTi dataset %(10,804,040 ratings, $S=\{1,\ldots,10\}$) 
and Amazon books\index{Amazon!Books Dataset}\index{Dataset!Amazon Books}. %(4,505,893 ratings, $S=\{1,\ldots,5\}$). 
The left column displays the accuracy of the algorithms in each dataset, i.e., the fraction of ratings that are exactly predicted by each algorithm.  The right column displays the mean absolute error (MAE) in the predicted vs.\ actual rating, treated as an integer or half-integer.  In all cases, the bars are the average of a five-fold cross-validation and the error bars correspond to the standard error of the mean.  The bipartite SBM algorithm does not scale to the larger datasets, hence it was evaluated only on the MovieLens\index{MovieLens} 100K and Yahoo Songs datasets. Importantly, bipartite SMB algorithm achieves similar accuracy to the MMSBM on the datasets it can handle.  The MMSBM model and algorithm achieves the best (highest) accuracy in five out of six datasets, and the best (lowest) MAE in four out of six datasets.
%Original source:  \protect\cite{guimera09}. \textit{Reprinted with permission.}.
}
\label{fig:compare}      
\end{figure}

To check the predictive power of the different algorithms we show the results for a five-fold cross validation in each of the six datasets. That is, we divide each dataset into five equal parts and we make the five possible combinations of using one part as the test set and the other four as the training set. We measure the predictability in terms of accuracy, that is the number of ratings predicted correctly, and the mean absolute error (MAE). Figure~\ref{fig:compare} shows the performance for the two network-based models, the simple bipartite SBM (using the approach in~\cite{guimera12}) and the Mixed-Membership Stochastic Block Model (MMSBM) (using the approach of~\cite{godoy-lorite16b}). Moreover, we show the comparison of the network-based approaches with three benchmark algorithms (see Sect.~\ref{sec:Benchmarcks}): the Item-Item\index{Item-Item!Algorithm}\index{Algorithm!Item-Item} algorithm \cite{sarwar01}, which predicts $r_{ui}$ based on the observed ratings of user $u$ for items that are the most similar to $i$, a ``classical'' Matrix Factorization\index{Matrix Factorization} (MF)~\cite{koren09}, and Mixed-Membership Matrix Factorization\index{Mixed-Membership Matrix Factorization}\index{Mixed-Membership!Matrix Factorization}\index{Matrix Factorization!Mixed-Membership} (MMMF) \cite{mackey10}; as well as a baseline naive algorithm that assigns to each test rating $r_{ui}$ the average of the observed ratings for item $i$, that is $ r_{ui} = \frac{1}{d_i} \sum_{u' \in \partial_i} r_{u'i} $. 
%For all these benchmark algorithms (except for the naïve model) we use the implementation in the LensKit package~\cite{ekstrand11}. 

The results for the MMSBM are for $K=L=10$, i.e. that is 10 groups of users and 10 groups of items (recall that there is any correspondence between these groups).  The performance for larger choices of $K$ and $L$ does not improve significantly \cite{godoy-lorite16b}. Since iterating the EM equation of Eqs.~\eqref{eq:update-omega}-\eqref{eq:update-pr} can lead to different solutions depending on the initial conditions, the results correspond to an average of the predicted probability distribution of ratings over 500 independent runs. There is a freely available implementation of the MMSMB in gitHub by Bill Jeffries (https://github.com/billjeffries/mixMemRec). The code is written in Spark's recommender library and can process large datasets. Notice however than the current implementation gives the results for a single run, therefore one should expect accuracies to be lower if a single run is considered.

The bipartite SBM does not require a pre-specification of model parameters since they are sampled by the algorithm. You can find a freely available implementation of the code in (http://seeslab.info/downloads/network-c-libraries-rgraph/ in rgraph-2.2.1/recommender/). For the Item-Item\index{Item-Item!Algorithm}\index{Algorithm!Item-Item} algorithm implemented in Lenskit, we set $k=50$; for the Matrix Factorization\index{Matrix Factorization} we also used the Lenskit implementation with $k=50$ features, a learning rate of $0.002$ and an initialization of $0.1$ for every user-feature and item-feature value as suggested in~\cite{ekstrand11}. Finally, for MMMF we use the Matlab\index{Matlab} implementation provided by the authors (https://code.google.com/archive/p/m3f/).

Another thing to take into account is that the network-based models, both the bipartite SBM and the MMSBM, are probabilistic models. That means that for each rating a user gives an item we have a probability distribution of ratings that results from the average of the probabilities for all the sampling set. Therefore, we can choose how to make predictions from the probability distribution of ratings: the mode (that is the rating with the highest probability), the mean or the median. As stated earlier, we measure the performance in terms of accuracy and the mean absolute error (MAE), which gives us an idea how far predictions are from the real values. For the network-based model, the best estimator for the accuracy is the most likely rating from the probability distribution of ratings, while for the MAE the best estimator is the median. In contrast, the predictions of the MF, MMMF and Item-Item\index{Item-Item!Model}\index{Model!Item-Item} models are a single real per rating.
 
In Fig.~\ref{fig:compare} we find that in most cases the network-based approaches, the bipartite SBM and MMSBM, outperform the Item-Item\index{Item-Item!Algorithm}\index{Algorithm!Item-Item} algorithm, MF and MMMF. Indeed, when considering the accuracy the MMSBM is significantly better than MF and MMMF for all the datasets we tested, and better than the item-item algorithm in five out of six datasets, the only exception being the Amazon Books dataset. In terms of the mean absolute error (MAE), the MMSBM is the most accurate in four out of the six datasets (item-item, MF and MMMF produce smaller MAE in the Amazon Books and MovieLens\index{MovieLens} 10M datasets). Note that the Amazon\index{Amazon} dataset is different from the others in that users only rate items after buying them, and knowing a priori the average rating of the item given by previous buyers, which might bias their choices. 

Interestingly, the MMSBM approach produces results that are almost identical to those of the bipartite SBM~\cite{guimera12} for the two examples for which inference with the bipartite SBM is feasible. 
In particular, the MMSBM achieves the same accuracy with $K=L=10$ in the mixed-membership model as the bipartite SBM with sampling models with $50$ groups on average. This suggests that many of the groups observed in~\cite{guimera12} are in fact mixtures of larger groups, and that the additional expressiveness of the MMSBM allows us to succeed with a lower-dimensional model.

Given that in general Matrix Factorization\index{Matrix Factorization} approaches outperform the Item-Item\index{Item-Item!Model}\index{Model!Item-Item} model, and that the MMSBM and the bipartite SMB give similar results, we  quantify the improvement of the MMSBM over the classical MF (with very similar results to the MMMF). To do so, we compute the relative improvements in the accuracy ($\%$) as
\begin{equation}
\frac{\rm(acc_{MMSBM})-\rm(acc_{MF})}{\rm(acc_{MF})}*100,
\end{equation} 
with improvements of $5\%$ in the MoviLens 100K dataset, $45\%$ in the MovieLens 10M, $41\%$ in the Yahoo songs, $42\%$ in the M-F LibimSeti agency, $27\%$ in the F-M LibimSeti agency and finally a $3\%$ improvement in the Amazon dataset. For the Amazon\index{Amazon} dataset, the relative improvement of the Item-Item\index{Item-Item!Algorithm}\index{Algorithm!Item-Item} over the MMSBM is of $13\%$.
%ML100K %.4279-0.4488%, ML 10M %0.27825811-0.40282589!!!!!!!!!%, yahoo %0.37755213-0.5312255!!!!!!!!!!%, MF %0.34982519-0.49733259!!!% FM %0.38187937-0.48597654!!!!% amazon %0.81511222-0.721152276% item Amazon %0.69702589-0.72115276%
%In absolut terms, the improvemnt of the network-based approaches with respect of the MF are:for the Movilens 100k we egt an improvemnt of $\sim 5\%$  , on the MovieLens 10M of $45\%$ , on the Yahoo songs $41\%$ , on the M-F Lenskit agency $42\%$  F-M $\sim 27\%$  and finaly Amazon dataset $3\%$  but in this case the item-item improves MMSBM in $13\%$.

%\section{Practical implementation of the network-based model MMSBM}
%\label{sec:example}
%In this section we will explain step by step how to run the MMSBM algorithm on a particular dataset, the MovieLens 100K dataset as an example. The code we are going to use is an implementation in Spark's recommender library by Bill Jefries (https://github.com/billjeffries/mixMemRec) freely available in git Hub. 

\section{Discussion}
\label{sec:discussion}

%-----------------------------------------------DEFFFF

We have shown that network-based approaches, based on inference using the block structure of social networks, give predictions of human preferences that are in most of the cases significantly and considerably more accurate than leading collaborative filtering recommendation algorithms.

In the case of the simple bipartite SBM, it is worth noting that the gain in accuracy comes at the expense of computational cost as a result of the Monte Carlo sampling of the user and item partition space. Although the algorithm is able to give predictions on datasets in the order of $\sim 1,000$ of users and items and $\sim 100,000$ of ratings, handling even one order of magnitude is challenging. Instead, the MMSBM inference using expectation-maximization method results in a scalable algorithm able to handle datasets with millions of ratings. 

In any case, network-based recommender systems not only provide better predictions, but also have some desirable features: they are analytically tractable allowing for a mathematically rigorous approach, they are based on plausible social models, and they provide interpretable results.

With respect to mathematical rigor, the Bayesian\index{Bayesian} approach used by the bipartite SBM \cite{guimera12} is the complete and correct probabilistic treatment of the observations. However, the results of the MMSBM suggest that introducing the mixed-membership of users and items is already equivalent to sampling over different sets of simple bipartite SBMs \cite{godoy-lorite16b}.

Importantly in both cases, we obtain an estimate of the whole probability distribution for each rating. From this, we can choose how to make predictions using the most likely rating, the mean or the median among others. In contrast, recommender systems like those based on Matrix Factorization\index{Matrix Factorization} or Item-Item\index{Item-Item!Algorithm}\index{Algorithm!Item-Item} give predictions that are a single number, the most likely rating (or a real number that should be rounded to the closer value in the ratings set), that may even be outside the rating range (for example, $r_{ui}=1.1$ when $S\in\{0,1\}$). Additionally, these algorithms assume that ratings are linearly spaced in the mind of users (that is, that the difference between $r=1$ and $r=2$ is the same as between $r=4$ and $r=5$), which does not seems to be in accordance with people's perception~\cite{ekstrand11}.

Finally, network-based approaches are based on models that were originally defined and are widely used to explain how social agents establish relationships, and is therefore in a better position to illuminate which social and psychological factors determine human preferences. As an interesting by-product of this, we note that it is possible to use them to infer demographic properties from ratings alone, a subject that is of much current interest for commercial purposes such as Social Marketing\index{Social Marketing}\index{Marketing!Social} \cite{leo16,andreasen02}.

The future of network-based recommender systems is likely to involve the introduction of contextual information about users and/or items into the inference process. The network-based models we have discussed in this chapter have the advantage that metadata can mathematically be introduced in the form of {\it a priori} probabilities for model parameters and specifically for group membership vectors. The intuition behind this idea is simple: we expect users (items) with similar associated metadata to have similar membership vectors. With this type of approach, network-based models will be better suited to industrially relevant problems such as the problem that arises when a new product is introduced into the market. The use of relevant metadata can be informative about the most plausible group membership vectors for each item and therefore help in identifying the range of users who could potentially be interested in that product.

%------------------------------------------------------------------------------------

%
%%
%%\section*{Appendix}
%%\addcontentsline{toc}{section}{Appendix}
%%%
%%%
%%When placed at the end of a chapter or contribution (as opposed to at the end of the book), the numbering of tables, figures, and equations in the appendix section continues on from that in the main text. Hence please \textit{do not} use the \verb|appendix| command when writing an appendix at the end of your chapter or contribution. If there is only one the appendix is designated ``Appendix'', or ``Appendix 1'', or ``Appendix 2'', etc. if there is more than one.
%%
%%\begin{equation}
%%a \times b = c
%%\end{equation}
%%
%\input{referenc}
\printindex
\end{document}